# Time-EAPCR-T: A Universal Deep Learning Approach for Anomaly Detection in Industrial Equipment

Huajie Liang[1,2], Di Wang[1,2], Yuchao Lu[1,2], Mengke Song[1,2], Lei Liu[1,2], Ling An[1,2], Ying Liang[1,2], Xingjie Ma[1,2], Zhenyu Zhang[1,2],*, Chichun Zhou[1,2],*

[1]School of Engineering, Dali University, Yunnan, 671003, China

[2]Air-Space-Ground Integrated Intelligence and Big Data Application Engineering Research Center of Yunnan Provincial Department of Education, Yunnan, 671003, China

*Corresponding author: Zhenyu Zhang: zhangzhenyu@dali.edu.cn

Chichun Zhou: zhouchichun@dali.edu.cn

Emails of other authors: Huajie Liang: lianghuajie@stu.dali.edu.cn

Di Wang: wangdi@stu.dali.edu.cn

Yuchao Lu: luyuchao@stu.dali.edu.cn

Mengke Song: songmenke@stu.dali.edu.cn

Lei Liu: liulei@stu.dali.edu.cn

Ling An: anling@stu.dali.edu.cn

Ying Liang: liangying@stu.dali.edu.cn

Xingjie Ma: maxingjie@stu.dali.edu.cn

# Abstract

With the advancement of Industry 4.0, intelligent manufacturing extensively employs sensors for real-time multidimensional data collection, playing a crucial role in equipment monitoring, process optimisation, and efficiency enhancement. Industrial data exhibit characteristics such as multi-source heterogeneity, nonlinearity, strong coupling, and temporal interactions, while also being affected by noise interference. These complexities make it challenging for traditional anomaly detection methods to extract key features, impacting detection accuracy and stability. Traditional machine learning approaches often struggle with such complex data due to limitations in processing capacity and generalisation ability, making them inadequate for practical applications. While deep learning feature extraction modules have demonstrated remarkable performance in image and text processing, they remain ineffective when applied to multi-source heterogeneous industrial data lacking explicit correlations. Moreover, existing multi-source heterogeneous data processing techniques still rely on dimensionality reduction and feature selection, which can lead to information loss and difficulty in capturing high-order interactions. To address these challenges, this study applies the EAPCR and Time-EAPCR models proposed in previous research and introduces a new model, Time-EAPCR-T, where Transformer replaces the LSTM module in the time-series processing component of Time-EAPCR. This modification effectively addresses multi-source data heterogeneity, facilitates efficient multi-source feature fusion, and enhances the temporal feature extraction capabilities of multi-source industrial data.Experimental results demonstrate that the proposed method outperforms existing approaches across four industrial datasets, highlighting its broad application potential.

# 1 Introduction

With the development of Industry 4.0 and the rapid advancement of intelligent manufacturing and industrial automation, a vast number of sensors are deployed in industrial environments to collect real-time, multidimensional data on equipment operating conditions and environmental parameters. These multi-source industrial data contain valuable information on equipment health status and production process efficiency. Effective analysis and utilisation of such data are crucial for condition monitoring, predictive maintenance, process optimisation, and product quality improvement.

In addressing industrial anomaly detection, Ajra et al. [1] proposed a fault detection and diagnosis (FDD) method integrating signal processing and intelligent computing for diagnosing critical components of VSI. However, this approach is affected by load variations, leading to a high false detection rate, increased computational complexity, and limited generalisation ability. Abdel Aziz et al. [2] reviewed machine learning-based prognostics and health management (PHM) techniques for engineering systems, discussing AI applications in condition monitoring and maintenance. However, machine learning relies on large-scale data and incurs high computational costs, making real-time implementation challenging. These limitations have driven researchers to explore more adaptive and intelligent solutions.

General-purpose large models have achieved groundbreaking progress in image and text domains, demonstrating exceptional adaptability and robustness across various tasks due to their powerful feature learning capabilities, generalisation ability, and efficient modelling of complex patterns. The industrial sector faces similar challenges, including high dimensionality, nonlinearity, and strong coupling. The success of general-purpose large models offers new perspectives for industrial intelligence, making them an inevitable trend in future industrial applications.

The development of large models relies significantly on deep learning models. Ochella et al. [3] systematically reviewed the applications of artificial intelligence (AI) techniques in prognostics and health management (PHM) for engineering systems, with a focus on the role of machine learning (ML) and deep learning (DL) models in condition monitoring and fault diagnosis. Studies indicate that deep learning models can efficiently process large-scale monitoring data, accurately identify potential fault patterns, and demonstrate superior performance in predictive maintenance.

With the increasing scale and rapid growth of industrial sensor data, the challenges of analysing multi-source, nonlinear, strongly coupled, and temporally interactive characteristics in industrial data have become more prominent. Traditional anomaly detection methods, such as rule-based threshold setting, expert judgment, and statistical analysis, exhibit low efficiency, high costs, and limited capability in accurately identifying complex anomaly patterns. While deep learning feature extraction modules, including CNN, RNN, and Transformer, have demonstrated outstanding performance in image and text data with explicit feature correlations, their effectiveness is limited when applied to multi-source industrial data lacking explicit feature associations.

During industrial equipment operation, state parameters are often generated in real time as multi-source heterogeneous data, exhibiting significant heterogeneity. Ensuring the safe and stable operation of industrial equipment requires extensive sensor data collection across multiple dimensions, such as temperature (°C), vibration (Hz), power (W), and current (A). These data not only differ in units but also vary in sampling frequency, sensor precision, and measurement range. Additionally, some data are continuous variables, such as temperature, current, and pressure, while others are discrete categorical variables, such as equipment operating states (low, medium, and high speed), further increasing the complexity of data fusion.

Additionally, equipment operating states typically change continuously over time, resulting in temporal interactions within the data. Moreover, multi-source sensor data exhibit cross-dimensional nonlinear temporal coupling, posing significant challenges for feature extraction and anomaly detection.

The heterogeneity and temporality of multi-source industrial data significantly impact feature extraction in several ways. First, challenges arising from data heterogeneity include: (1) Scale discrepancy: Differences in measurement units across sensors may lead to overemphasis on certain features while neglecting others; (2) Information loss: Improper data transformation or fusion can result in the loss of critical information, reducing anomaly detection accuracy; (3) Feature fusion complexity: Effectively handling heterogeneity to ensure no information loss during feature fusion is essential for improving model accuracy. Additionally, the temporal nature of the data presents further challenges: the relationships between multi-source data dynamically evolve over time. Capturing and integrating these temporal features while preventing information loss is crucial for enhancing model performance.

Existing methods for multi-source heterogeneous data feature extraction still exhibit several limitations in addressing the aforementioned challenges: (1) Dimensionality reduction techniques, such as PCA, t-SNE, and UMAP, fundamentally compress data, which may result in the loss of critical information, particularly for nonlinear features or high-dimensional complex relationships. Moreover, these methods typically rely on the global data structure and struggle to adaptively adjust when handling dynamic data or varying combinations of data sources. (2) Feature selection methods, including chi-square tests, mutual information, and L1 regularisation, primarily focus on the relationship between individual variables and the target variable, making them ineffective in capturing high-order interactions between multiple features. Furthermore, many feature selection techniques depend on statistical assumptions (e.g., chi-square tests assume variable independence), which can lead to performance degradation when the data distribution does not conform to these assumptions. (3) Deep learning methods, such as CNN, RNN, and Transformer, are well-suited for processing image, text, and numerical data with explicit feature correlations, often requiring specialised network structures for different data types. However, in multi-source data lacking explicit feature associations, effectively extracting key features remains a critical challenge.In summary, while existing approaches are effective in specific scenarios, they face considerable challenges in handling data heterogeneity, complex feature relationships, temporal interactions, and computational costs. Therefore, there is an urgent need for a unified feature extraction module capable of modelling multi-source heterogeneous industrial data effectively.

This study applies the EAPCR model [4] and Time-EAPCR model [5] to address anomaly detection in the industrial domain, introducing modifications to the time-series processing component of Time-EAPCR. Specifically, Transformer replaces the original LSTM-based time-series module, and the new model is named Time-EAPCR-T. Leveraging the self-attention mechanism, Transformer captures feature correlations globally, effectively modelling temporal interactions within multi-source industrial data. Combined with the multi-stage feature extraction and fusion training mechanism of the EAPCR model, the proposed approach comprehensively captures feature patterns of multi-source sensor data at the same time step. This model design significantly enhances detection accuracy and generalisation ability in industrial anomaly detection. Experimental results on real-world industrial datasets demonstrate that the proposed method outperforms existing anomaly detection approaches for multi-source heterogeneous industrial data.

## 2 Materials and methods

### 2.1 Dataset

To evaluate the effectiveness of the Time-EAPCR-T method in industrial anomaly detection, we utilise four publicly available datasets. These datasets are sourced from published studies and can be accessed through the references. A summary of the datasets is presented in Table 1.

EngineFaultDB [6]: This dataset is used for complex fault diagnosis of automobile engines. The data were collected from the widely used C14NE spark-ignition engine, comprising 55,999 records. The dataset includes 16,000 normal samples, 10,998 samples with rich mixture faults, 15,000 samples with lean mixture faults, and 14,001 samples with low voltage faults. Each record contains 14 variables, including Manifold Absolute Pressure (MAP), Throttle Position Sensor (TPS), and Force. In this study, the dataset is split into a training set (44,799 samples) and a test set (11,200 samples) using an 8:2 ratio.

Mendeley [7]: This dataset consists of intershaft bearing fault data from aero engines, collected using two eddy current sensors and four accelerometers at a sampling frequency of 25 kHz. The data were collected five times, covering 28 different rotational speed conditions, with each recording lasting 15 seconds. The raw data were segmented into 18 groups of samples, each with a length of 20,480, containing seven variables (displacement, acceleration, and rotational speed signals). The dataset includes three operating conditions: normal operation, inner ring fault, and outer ring fault. After removing invalid data, 2,412 sample groups remained. In this study, the train-test split follows the original dataset settings, with each category divided using a 7:3 ratio, resulting in a training set (34,578,432 samples) and a test set (14,819,328 samples).

SEU [8]: This dataset is used for gearbox fault diagnosis, containing data from bearings and gears. Each category includes two operating conditions and five working states, with approximately 1,048,560 samples per state. The dataset covers normal operation as well as various typical fault conditions, such as roller faults, inner ring faults, and outer ring faults. Each sample consists of eight variables, including vibration and torque signals. In this study, 10,000 samples were randomly selected from each category, resulting in a total of 200,000 samples. The dataset was then split into a training set (160,000 samples) and a test set (40,000 samples) using an 8:2 ratio, ensuring no overlap between the training and test sets.

**Table 1** The main features of datasets

| Dataset | Train | Test | Dimension | Number of Classes |
|---|---|---|---|---|
| EngineFaultDB | 44,799 | 11,200 | 14 | 4 |
| Mendeley | 34,578,432 | 14,819,328 | 7 | 3 |
| SEU | 160,000 | 40,000 | 8 | 20 |
| Gearbox Fault Diagnosis | 480,000 | 120,000 | 4 | 2 |

Gearbox Fault Diagnosis [8]: This dataset is used for gearbox fault diagnosis and was collected using vibration sensors positioned in four directions. The operating conditions cover load levels ranging from 0% to 90% in 10% increments, with approximately 88,000 samples per condition and state. The dataset consists of two categories: healthy condition and tooth breakage fault. Each sample contains four variables. In this study, 30,000 consecutive samples were randomly selected from each condition, resulting in a total of 600,000 samples. The dataset was then split into a training set (480,000 samples) and a test set (120,000 samples) using an 8:2 ratio, ensuring no overlap between the training and test sets.

## 2.2 Methodology

### 2.2.1 A review of EAPCR

The overall structure of the EAPCR model [4] is illustrated in Figure 1. This model effectively extracts features and patterns from multi-source heterogeneous data, primarily due to its unique network design. The following sections provide a detailed explanation of the functions of each component.

E (Embedding): The Embedding layer encodes individual features, mapping multi-source heterogeneous data into a unified feature space. This mapping not only addresses data heterogeneity but also establishes a foundation for subsequent heterogeneous feature fusion.

A (Attention): The Bilinear-attention mechanism computes the Gram matrix to capture feature similarities, thereby reflecting the relationships between features. This design facilitates the rapid extraction of effective features and enhances the model's understanding of feature interactions.

P (Permutation): The Permutation layer rearranges the Gram matrix to generate the G'matrix, expanding the range of CNN feature sampling in the next stage. This enhances the model's ability to learn diverse features, thereby improving its robustness.

C (CNN): The CNN module not only learns local correlations between features but also effectively captures complex and high-order interactions. Additionally, by leveraging the rearranged Gram matrix, CNN further expands the sampling range, enabling the extraction of long-range dependencies between features.

R(Residual): The model captures both local and long-range correlations between features through CNN while leveraging a Multi-Layer Perceptron (MLP) to extract global information from feature embeddings, addressing CNN's limitations in global feature extraction. By incorporating a residual structure, the model effectively integrates different feature extraction approaches, enhancing feature comprehensiveness and significantly improving its generalisation ability.

Through the synergistic interaction of these components, the EAPCR model achieves efficient feature extraction and fusion for multi-source heterogeneous data, providing strong support for subsequent anomaly detection tasks.

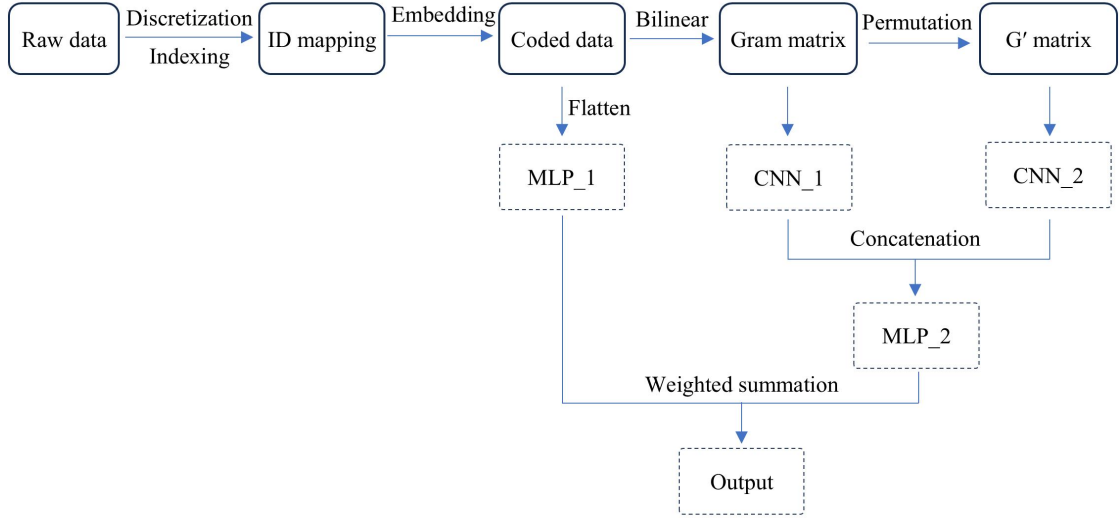

**Figure 1.** Structure of the EAPCR Model

## 2.2.2 Time-EAPCR

The Time-EAPCR model consists of a multi-sensor feature processing module and a time-series processing module. The multi-sensor feature processing module utilises EAPCR to extract correlation patterns among multi-source features. The time-series processing module employs an enhanced EAPCR to capture the temporal interactions of multi-source features over time.

The time-series processing module employs an enhanced EAPCR to capture temporal interactions among multi-source features. Unlike the discretization process in EAPCR, TAPCR applies ID encoding only when heterogeneous data contain categorical variables. This improvement reduces dependency on data discretization, decreases the number of hyperparameters requiring tuning, and thereby simplifies the model's workflow, enhancing its practicality and usability.

In this study, Transformer replaces the LSTM module in the time-series processing component of Time-EAPCR, which is referred to as TAPCR. The overall structure is illustrated in Figure 2-b, and the resulting model is named Time-EAPCR-T. In time-series feature extraction, LSTM accumulates local information gradually, making it well-suited for capturing short-range dependencies and performing effectively on stationary sequences. However, as LSTM relies on hidden state propagation for transmitting long-range information, it may suffer from information decay when processing long-distance dependencies. In contrast, Transformer, leveraging its self-attention mechanism, directly models the relationships between any two features within a time window, enabling effective capture of long-range dependencies. This makes Transformer particularly advantageous for handling complex patterns and multi-scale feature extraction tasks, especially for non-stationary or highly dynamic time-series data. The introduction of

Transformer enhances the model's ability to capture temporal interactions within multi-source industrial data, significantly improving anomaly detection accuracy and robustness.

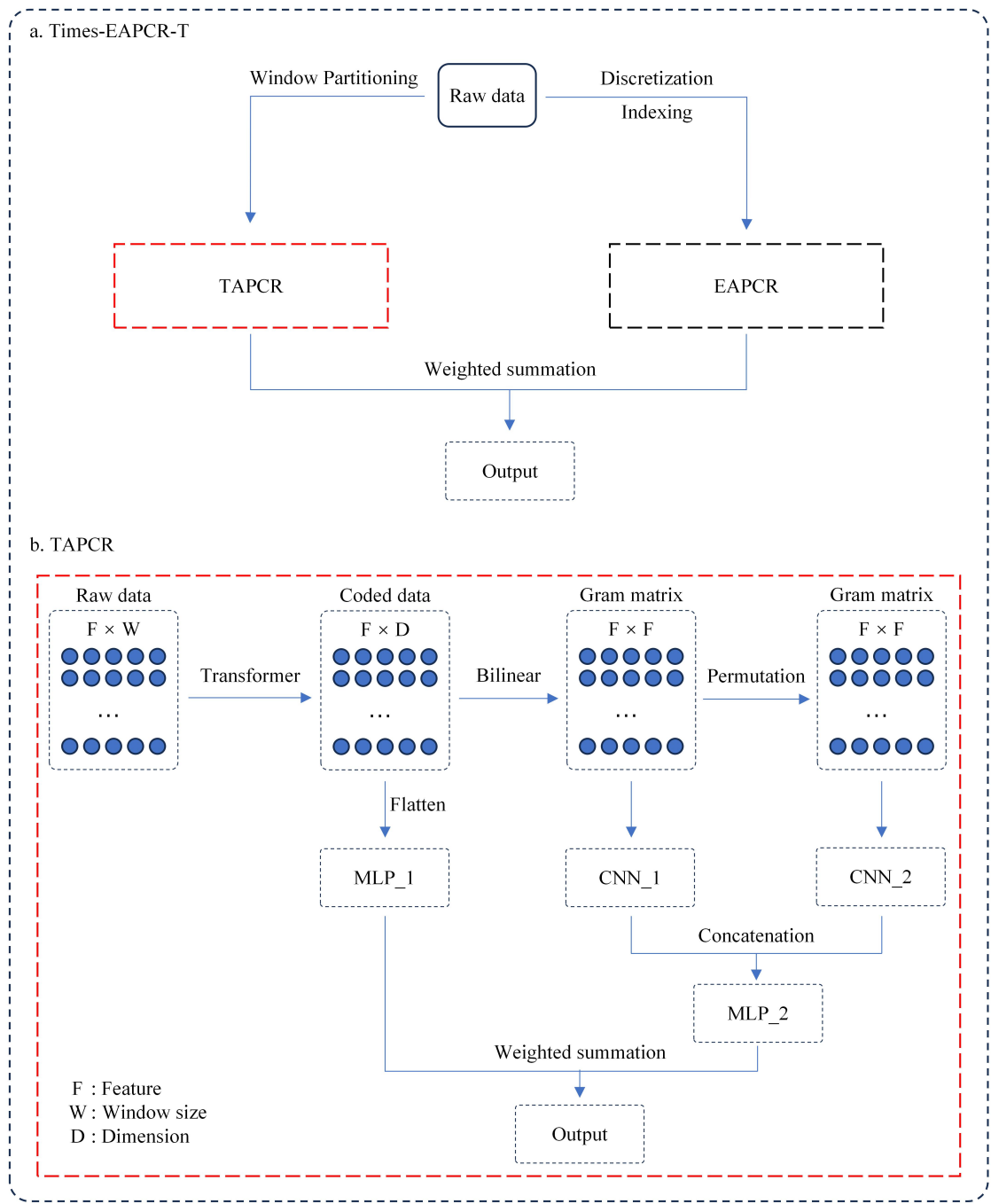

**Figure 2.** Structure of the Time-EAPCR-T and TAPCR Models

Time-EAPCR-T utilises a weighted summation mechanism to regulate the final output encoding, enabling the model to effectively capture both feature correlations within samples and temporal trends. This allows Time-EAPCR-T to be applicable to datasets with or without temporal dependencies, enhancing its generality and adaptability in processing multi-source heterogeneous industrial data. As a result, it provides a more flexible and efficient solution for industrial equipment anomaly detection.

**Table 2** Network Architecture of Time-EAPCR-T

| Modules | Layers | Input Dimension | Output Dimension | Parameters |
|---|---|---|---|---|
| Embedding | Embedding | (Batch, Steps, 128) | (Batch, Steps, 128) | |
| MLP_1 | Linear_1 | (Batch, Steps*128) | (Batch, Steps*32) | |
| | Dropout | | | p = 0.5 |
| | Linear_2 | (Batch, Steps*32) | (Batch, Output_size) | |
| MLP_2 | Linear_1 | (Batch, Steps*128) | (Batch, Steps*32) | |
| | Dropout | | | p = 0.5 |
| | Linear_2 | (Batch, Steps*32) | (Batch, Output_size) | |
| CNN_1 | Conv2d_1 | (Batch, 1, Steps, Steps) | (Batch, 4, Steps, Steps) | Kernel=5 padding=2 |
| | Conv2d_2 | (Batch, 4, Steps, Steps) | (Batch, 2, Steps, Steps) | Kernel=3 padding=1 |
| CNN_2 | Conv2d_1 | (Batch, 1, Steps, Steps) | (Batch, 4, Steps, Steps) | Kernel=5 padding=2 |
| | Conv2d_2 | (Batch, 4, Steps, Steps) | (Batch, 2, Steps, Steps) | Kernel=3 padding=1 |

### 2.2.3 Evaluation Metrics

In this study, to evaluate the classification performance of the Time-EAPCR-T method, we adopt Accuracy, Recall, and F1-score as evaluation metrics. These metrics are widely used in multi-class classification tasks and effectively assess the overall predictive performance of the model. In particular, they provide a more comprehensive evaluation when dealing with imbalanced class distributions. Before introducing the specific metrics, we first define the four fundamental concepts in multi-class classification tasks.

True Positive (TP): The number of samples from class k that are correctly classified as k. If $CM_{k,k}$ represents the number of samples in the confusion matrix where the true class is k and the predicted class is also k, it can be expressed mathematically as:

$$TP_k = CM_{k,k} \qquad (1)$$

False Positive (FP): The number of samples predicted as class k by the model, but whose actual class is not k. This is calculated as the sum of all elements in column k of the confusion matrix (i.e., the total number of samples predicted as k), minus the diagonal element $TP_k$. The mathematical expression is given by:

$$FP_k = \sum_{i=0}^{K-1} CM_{i,k} - TP_k \qquad (2)$$

True Negative (TN): The number of samples that are neither class k nor predicted as class k. This is calculated as the sum of all elements in the confusion matrix, minus the TP, FP, and FN of class k. The mathematical expression is given by:

$$TN_k = \sum_{i=0}^{K-1}\sum_{j=0}^{K-1} CM_{i,j} - (TP_k + FP_k + FN_k) \qquad (4)$$

Accuracy: Accuracy measures the proportion of correctly classified samples among all samples. The numerator represents the sum of all true positives across all classes (i.e., the sum of the diagonal elements of the confusion matrix), while the denominator is the total number of samples. The mathematical expression is given by:

$$Accuracy = \frac{\sum_{k=0}^{K-1} TP_k}{\sum_{i=0}^{K-1}\sum_{j=0}^{K-1} CM_{i,j}} \qquad (5)$$

Precision: Precision measures the proportion of correctly predicted samples for a given class among all samples predicted as that class. Since the Macro F1-score is calculated in the following sections, this refers to Macro Precision, denoted as Precision. The specific formula is:

$$Precision = \frac{1}{K}\sum_{k=0}^{K-1} \frac{TP_k}{TP_k + FP_k} \qquad (6)$$

Recall: Recall measures the proportion of correctly identified samples within a given class. In multi-class classification tasks, Macro Recall (denoted as Recall) is commonly used. This metric calculates recall for each class individually and then averages the values, ensuring equal weight for all classes. The specific formula is:

$$Recall = \frac{1}{K}\sum_{k=0}^{K-1} \frac{TP_k}{TP_k + FN_k} \qquad (7)$$

F1-score: The F1-score is the harmonic mean of Precision and Recall. In cases of class imbalance, the F1-score provides a more reliable evaluation than Accuracy. This study adopts the Macro F1-score, denoted as F1, which is calculated as follows:

$$F1 = \frac{2 \times M\_Precision \times M\_Recall}{M\_Precision + M\_Recall} \qquad (8)$$

# 3 Result and analysis

## 3.1 Result

This study evaluates the performance of the Time-EAPCR-T model on four publicly available datasets (EngineFaultDB, SEU, Gearbox Fault Diagnosis, and Mendeley). The same model architecture is applied across all datasets, with no parameter adjustments specific to any dataset. The detailed model parameters are presented in Table 2, and the experimental results are shown in Table 3.

**Table 3** Main result

| Datasets | Methods | F1_score | Accuracy | Recall |
|---|---|---|---|---|
| Mendeley [7] | CNN | / | 0.831 | / |
| | LSTM | / | 0.854 | / |
| | TST | / | 0.711 | / |
| | **Time-EAPCR-T** | **0.982** | **0.980** | **0.980** |
| EngineFaultDB [9] | Logistic Regression | 0.574 | 0.576 | 0.576 |
| | Decision Tree | 0.750 | 0.750 | 0.750 |
| | Random Forest | 0.748 | 0.748 | 0.748 |
| | SVC | 0.715 | 0.747 | 0.747 |
| | KNN | 0.751 | 0.751 | 0.751 |
| | Naïve Bayes | 0.353 | 0.394 | 0.394 |
| | Neural Net. | 0.748 | 0.749 | 0.749 |
| | **Time-EAPCR-T** | **0.760** | **0.754** | **0.760** |
| SEU [10] | Resnet18-1d (2016) | 0.8608 | 0.963 | 0.873 |
| | WDCNN (2017) | 0.850 | 0.740 | 0.854 |
| | QCNN (2022) | 0.850 | 0.966 | 0.863 |
| | WKNet1_Inception (2022) | 0.929 | 0.960 | 0.922 |
| | MRA-CNN (2022) | 0.925 | 0.951 | 0.938 |
| | MFRANet | 1.000 | 0.988 | 1.000 |
| | **Time-EAPCR-T** | **1.000** | **0.995** | **1.000** |
| Gearbox Fault Diagnosis [11] | XGBClassifier | 0.779 | 0.782 | 0.752 |
| | KNeighbors Classifier | 0.662 | 0.637 | 0.662 |
| | ExtraTrees Classifier | 0.924 | 0.927 | 0.902 |
| | Decision Tree Classifier | 0.780 | 0.751 | 0.795 |
| | Gradient Boosting Classifier | 0.943 | 0.953 | 0.928 |
| | MLP Classifier | 0.651 | 0.653 | 0.663 |
| | Linear SVC | 0.502 | 0.542 | 0.521 |
| | SGD Classifier | 0.919 | 0.926 | 0.913 |

| | | | |
|---|---|---|---|
| Random Forest Classifier | 0.654 | 0.663 | 0.685 |
| Bernoul liNB | 0.923 | 0.920 | 0.894 |
| Proposed model | 0.962 | 0.986 | 0.923 |
| **Time-EAPCR-T** | **1.000** | **1.000** | **1.000** |

EngineFaultDB does not contain temporal dependencies, whereas the other three datasets incorporate temporal relationships. Experimental results demonstrate that Time-EAPCR-T outperforms existing methods across all datasets. To ensure the stability and reproducibility of the results, all performance metrics are reported as the average values over the last 20 epochs on the test set. The detailed experimental results are presented in Table 3, where the "Other Methods" column displays the best-reported model results from relevant literature for each dataset.

In the SEU dataset, the Accuracy of other methods corresponds to the 20-class classification task, while F1-score and Recall are reported for the 15-class classification task (refer to the original paper's confusion matrix section for details). To ensure consistency with other methods, F1-score and Recall in Table 3 are also reported for the 15-class classification task. For the 20-class classification task, the values of F1-score, Accuracy, and Recall are all 0.995.

As shown in Table 3, Time-EAPCR-T achieves outstanding performance across all four public datasets. In EngineFaultDB, Time-EAPCR-T outperforms other methods in terms of F1-score, Accuracy, and Recall, demonstrating its effectiveness in handling anomaly detection tasks without temporal dependencies.

In datasets with temporal dependencies, the F1-score and Recall of Time-EAPCR-T on the SEU dataset are equivalent to those of other methods, while its Accuracy is slightly higher. On the Gearbox Fault Diagnosis and Mendeley datasets, Time-EAPCR-T outperforms other methods across all metrics, demonstrating its strong capability in modeling multi-source heterogeneous sequences.

In summary, the Time-EAPCR-T model demonstrates exceptional robustness and anomaly detection capability across different datasets. It performs particularly well on datasets without temporal dependencies while also effectively adapting to time-series data, enabling accurate detection through the extraction of key features.

## 3.2 Analysis

The primary hyperparameter of the Time-EAPCR-T model is the Window Size in TAPCR, which significantly affects the model's performance. In Table 3, the EngineFaultDB dataset does not contain temporal dependencies, meaning that features are independent across the time dimension rather than forming continuous sequences. Therefore, the Window Size for this dataset is set to 1. For datasets with temporal dependencies, the Window Sizes for SEU (15-class classification), Gearbox Fault Diagnosis, and Mendeley are set to 256, 1024, and 32, respectively. To further analyse the impact of Window Size, the performance of Time-EAPCR-T on these three datasets under different Window Size settings is presented in Table 4. The results indicate that Time-EAPCR-T achieves optimal performance across all three datasets.

Table 4. Differences in M_F1 Values of Time-EAPCR-T Across Different Window Sizes

| Datasets | Window sizes | | | | | | |
|---|---|---|---|---|---|---|---|
| | 16 | 32 | 64 | 128 | 256 | 512 | 1024 |
| SEU (15-Class Classification) | 0.947 | 0.980 | 0.994 | 0.997 | **1.000** | **1.000** | **1.000** |
| Gearbox Fault Diagnosis | 0.998 | **1.000** | **1.000** | **1.000** | **1.000** | **1.000** | **1.000** |
| Mendeley | 0.926 | 0.928 | 0.915 | 0.958 | 0.980 | **0.983** | 0.959 |

### 3.3 Ablation Study

To further validate the effectiveness of each component in the Time-EAPCR-T model, this section presents an ablation study. We conducted comparative experiments on the same datasets to evaluate the contributions of the TAPCR module, EAPCR module, and the full Time-EAPCR-T model. The experimental results, shown in Table 5, indicate that the highest values are highlighted in bold, while the second-highest values are underlined.

Table 5. Ablation Study Results of the Time-EAPCR-T Model

| Datasets | Methods | M_F1 | Accuracy | M_Recall |
|---|---|---|---|---|
| EngineFaultDB | EAPCR | 0.759 | 0.753 | <u>0.759</u> |
| | TAPCR | 0.735 | <u>0.751</u> | 0.757 |
| | Time-EAPCR | <u>0.758</u> | <u>0.751</u> | 0.758 |
| | Time-EAPCR-T | **0.760** | **0.754** | **0.760** |
| SEU (15-Class Classification) | EAPCR | 0.477 | 0.485 | 0.485 |
| | TAPCR | <u>0.999</u> | <u>0.999</u> | <u>0.999</u> |
| | Time-EAPCR | 0.992 | 0.992 | 0.992 |
| | Time-EAPCR-T | **1.000** | **1.000** | **1.000** |
| Gearbox Fault Diagnosis | EAPCR | <u>0.606</u> | <u>0.611</u> | <u>0.611</u> |
| | TAPCR | **1.000** | **1.000** | **1.000** |
| | Time-EAPCR | **1.000** | **1.000** | **1.000** |
| | Time-EAPCR-T | **1.000** | **1.000** | **1.000** |
| Mendeley | EAPCR | 0.562 | 0.630 | 0.561 |
| | TAPCR | 0.965 | 0.967 | 0.964 |

|  |  |  |  |
|---|---|---|---|
| Time-EAPCR | **1.000** | **1.000** | **1.000** |
| Time-EAPCR-T | 0.980 | 0.982 | 0.980 |

The experimental results indicate that on the Mendeley dataset, the Time-EAPCR-T model performs slightly worse than other models. Meanwhile, on the Gearbox Fault Diagnosis dataset, Time-EAPCR-T, TAPCR, and Time-EAPCR achieve identical performance, with 100% Accuracy and Recall in anomaly detection. However, from a comprehensive evaluation perspective, the Time-EAPCR-T model demonstrates greater stability compared to other models.

# 4 Conclusion and Outlook

## 4.1 Conclusion

This study presents an improved Time-EAPCR-T model, a novel approach for industrial anomaly detection that integrates the advantages of TAPCR and EAPCR to address the challenges of multi-source heterogeneous data. The main contributions of this study are summarised as follows:

1) Enhanced Time-EAPCR model: The proposed model demonstrates outstanding performance in multi-source industrial environments, outperforming the best existing methods on five public datasets, validating its potential as a general anomaly detection model.

2) Introduction of the Transformer module and weighted summation mechanism: By replacing the Embedding module with Transformer, Time-EAPCR effectively captures temporal dependencies in time-series data. Additionally, the weighted summation mechanism enables the model to simultaneously capture feature-level and trend-level correlations, further enhancing its overall performance.

## 4.2 Outlook

Although Time-EAPCR-T demonstrates excellent performance in multi-source data, effectively capturing temporal features and sample correlations, several aspects warrant further exploration:

1) Optimisation of noise robustness: While the current model performs well, there is room for improvement when handling high-noise datasets (e.g., EngineFaultDB). Future research could develop more robust feature extraction methods or noise filtering mechanisms, such as wavelet transforms or adaptive filtering techniques, to enhance the model's stability in complex industrial environments.

2) Enhancing computational efficiency: As industrial data continues to grow in scale, computational efficiency becomes a critical challenge. Future work could explore lightweight architectures (e.g., depthwise separable convolutions) or distributed computing frameworks to improve the model's ability to process high-dimensional data, making it more suitable for large-scale industrial applications.

3) Cross-domain adaptability validation: While Time-EAPCR-T has demonstrated strong performance in industrial anomaly detection, its applicability in other domains, such as medical diagnostics and financial risk assessment, remains unverified. Future research could extend its application to diverse fields to evaluate its generalisability and scalability.

In conclusion, Time-EAPCR provides a powerful and flexible solution for industrial anomaly detection, offering significant improvements over existing methods. We believe this study lays a solid foundation for future research on multi-source heterogeneous data analysis and opens up new possibilities for practical applications across various domains.

## Data and Code Availability

The public datasets can be found in the corresponding references. The source code and private dataset can be made available upon reasonable request to the corresponding author.

## Acknowledgments

This work was supported by the National Natural Science Foundation of China (42367066, 62106033), Yunnan Fundamental Research Projects (202401AT070016, 202301BA070001-037), and National Observation and Research Station of Erhai Lake Ecosystem in Yunnan (2022ZZ01)

## Author contribution statement

Conceptualization, Chichun Zhou and Zhenyu Zhang; Experimental work, Huajie Liang, Xingjie Ma, Di Wang; Formal analysis, Ling An, Chichun Zhou, Lei Liu, Yuchao Lu; Investigation, Di Wang, Xingjie Ma, Ying Liang, Ling An, Ying Liang; Writing - original draft, Huajie Liang, Ying Liang, Yuchao Lu, Mengke Song; Funding acquisition, Chichun Zhou, Zhenyu Zhang; Writing-review & editing, Chichun Zhou, Zhenyu Zhang and Ling An; All the authors have read and agreed to the published version of the manuscript.

## Additional information

Conflicts of Interest: The authors declare that they have no known competing financial interests or personal relationships that could have appeared to influence the work reported in this paper.